\documentclass[10pt,twocolumn,letterpaper]{article}
\usepackage[]{cvpr}
\usepackage{microtype}

\usepackage{xspace}
\newcommand{\methodname}{ChangeWAM\xspace}

\AtBeginDocument{%
  \setlength{\abovedisplayskip}{7.5pt plus 2pt minus 2pt}%
  \setlength{\belowdisplayskip}{7.5pt plus 2pt minus 2pt}%
  \setlength{\abovedisplayshortskip}{0pt plus 2pt}%
  \setlength{\belowdisplayshortskip}{5pt plus 2pt minus 2pt}%
}

\definecolor{cvprblue}{rgb}{0.21,0.49,0.74}
\usepackage[pagebackref,breaklinks,colorlinks,allcolors=cvprblue]{hyperref}
\usepackage{multirow}
\def\confName{CVPR}
\def\confYear{2027}
\title{\methodname: Bidirectional Change Guidance for World–Action Models}

\author{
Hanxue Zhang$^{1,2}$,
Baole Fang$^{2}$,
Yating Wang$^{2,3}$,
Mulin Yu$^{2}$,
Tao Lu$^{2}$ \\
Junting Dong$^{2}$,
Chunhua Shen$^{2,4}$,
Qingsong Yao$^{5*}$,
Haoyu Guo$^{2*}$\\
[2mm]
$^1$~Shanghai Jiao Tong University \quad
$^2$~Shanghai AI Laboratory \\
$^3$~Tongji University \quad
$^4$~Zhejiang University  \quad
$^5$~Renmin University of China 
\\
[1mm]
}

\newif\ifwithappendix
\withappendixtrue

\begin{document}

\maketitle

{
\let\thefootnote \relax 
\footnote{$^\ast$Qingsong Yao and Haoyu Guo are corresponding authors. }
}

\begin{abstract}
World-action models (WAMs) integrate future-video prediction and action generation to improve robotic manipulation. 
However, how to effectively organize the interaction between world and action experts remains an important question. 
We propose \methodname, a framework that establishes bidirectional guidance through change prediction.
The video expert predicts how the scene will change to guide action generation, while the action expert predicts the robot motion implied by its current action hypothesis to guide future-video prediction.
We realize this interaction through visual and depth change prediction together with robot-motion supervision, grounding the exchanged representations in manipulation-relevant changes.
To further strengthen this bidirectional guidance, we introduce feature alignment and feedback, allowing predictive knowledge acquired through action-conditioned world learning to benefit action generation without requiring future-video generation at inference. 
Experiments on three out-of-distribution benchmarks demonstrate the effectiveness of \methodname as an add-on to multiple WAMs.
Incorporating our proposed mechanisms into 4D-WAM improves success rates by 11.13, 3.93, and 18.76 percentage points on LIBERO-Plus, LIBERO-Pro, and RoboTwin 2.0 clean-to-random, respectively.
These results highlight the effectiveness of change-based bidirectional guidance for generalizable robotic manipulation.
Code and model checkpoints will be publicly released.
\end{abstract}
\section{Introduction}
\label{sec:intro}

\begin{figure}[t]
    \centering
    \vspace{-4mm}
    \includegraphics[width=0.96\linewidth]{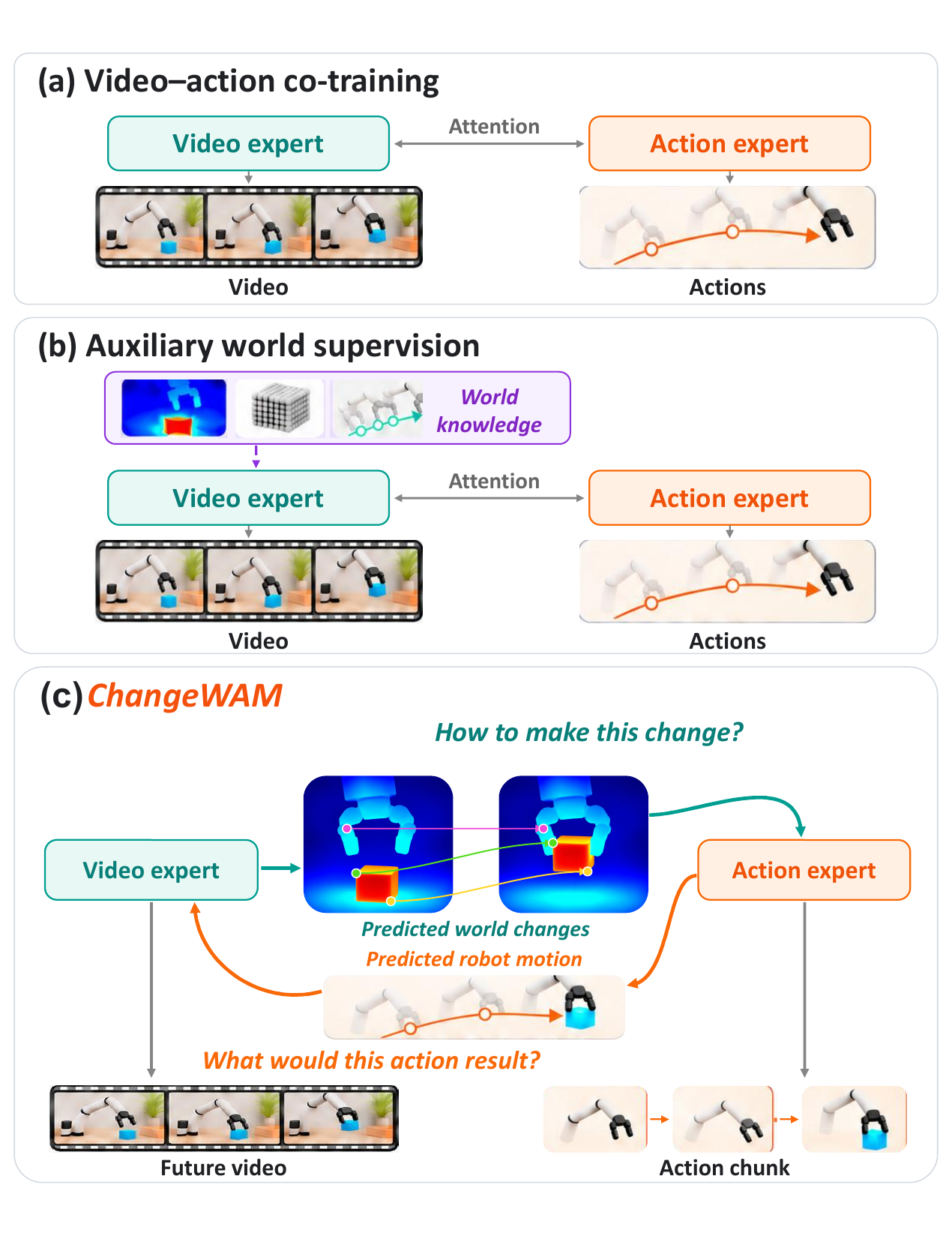}
    \vspace{-4mm}
    \caption{
      \textbf{Change-based bidirectional guidance for world--action models.}
      (a) WAMs couple video prediction and action generation through attention~\cite{lingbotva,fastwam}.
      (b) Auxiliary world supervision enriches this learning with additional geometric and motion knowledge~\cite{4dwam,dreamwam}.
      (c) \methodname connects the two experts through complementary predictions:
      anticipated world changes guide action generation (\emph{How to make this change?}),
      while the robot motion predicted from candidate actions guides future-world prediction
      (\emph{What would this action result?}).
      }
    \label{fig:teaser}
    \vspace{-1mm}
\end{figure}

World-action models (WAMs) bring visual prediction and action generation into a common learning framework, offering a way to ground robot control in predictive knowledge of the physical world \cite{lingbotva,fastwam,internw0delta}. Learning how scenes evolve provides supervision about object motion, spatial relationships, and interaction dynamics beyond action labels alone. Coupling this supervision with policy learning offers a route to use advances in video modeling for robotic manipulation. WAMs therefore provide a promising framework for learning robot behavior through both visual dynamics and action supervision.

Recent work has strengthened WAMs through architectures that coordinate world and action experts \cite{lingbotva,fastwam,wsa1,internw0delta} (\Cref{fig:teaser}(a)), and through incorporating geometric, motion, and semantic priors into world-action learning \cite{4dwam,dreamwam,wsa1} (\Cref{fig:teaser}(b)).
We highlight \emph{change prediction} as a shared basis for mutual guidance between world and action experts.
Manipulation fundamentally involves changing the physical world, making predicted changes a natural link between understanding how the world may evolve and deciding how to act. 
Change prediction also provides a more focused learning signal for manipulation, as it emphasizes evolving geometry and motion while placing less emphasis on persistent appearance details such as background texture.

To realize this idea, we propose \methodname, a change-based bidirectional framework for world-action models.
As illustrated in \Cref{fig:teaser}(c), \methodname connects two complementary predictive relationships between world changes and robot actions.
In one direction, anticipating how the world will change provides predictive guidance for action generation. 
In the other, how the robot would move under candidate actions suggests possible scene changes, which can in turn guide future-world prediction.
By integrating these complementary predictive relationships into joint learning, \methodname enables world prediction and action generation to inform and improve one another.

Specifically, we introduce additional world-change tokens in the video expert to predict video differences, depth differences, and end-effector motion. 
The action expert reads these tokens through attention to guide action generation. 
On the action side, we introduce action-effect tokens that read the current action hypothesis and predict the motion it implies, providing guidance for future-video prediction through cross-attention.
We further introduce a predictive feature alignment and feedback mechanism. 
During training, we align the action-effect tokens with the video expert’s predictive features. We feed these tokens back into the action expert through attention to improve action generation. 
This design allows the action expert to benefit from action-guided video prediction without generating future video at inference.

Experiments on three out-of-distribution benchmarks demonstrate that the mechanisms proposed in \methodname improve multiple base WAMs.
When incorporated into 4D-WAM, these mechanisms increase success rates by 11.13, 3.93, and 18.76 percentage points on LIBERO-Plus, LIBERO-Pro, and RoboTwin 2.0 clean-to-random, respectively.
On LIBERO-Plus, our approach also outperforms the compared methods that enhance WAMs through additional spatial and motion supervision.
These results highlight the potential of using change-based bidirectional guidance to make world prediction and action generation support one another in learning generalizable robot behavior.

Our contributions are as follows:
\begin{itemize}
    \item We propose \methodname, a framework that connects world prediction and action generation through change-based bidirectional guidance.
    \item We introduce world-change tokens supervised by video differences, depth differences, and end-effector motion, providing manipulation-related predictive features to improve action generation.
    \item We introduce action-effect tokens that predict robot motion to improve future-video prediction, and develop predictive feature alignment and feedback to bring the resulting video-side knowledge back to action generation.
    \item We demonstrate consistent improvements by applying our proposed mechanisms into multiple WAMs on three out-of-distribution manipulation benchmarks, achieving success-rate gains of 11.13, 3.93, and 18.76 percentage points on LIBERO-Plus, LIBERO-Pro, and RoboTwin 2.0 clean-to-random,  respectively.
\end{itemize}

\begin{figure*}[t]
    \centering
    \includegraphics[width=\textwidth]{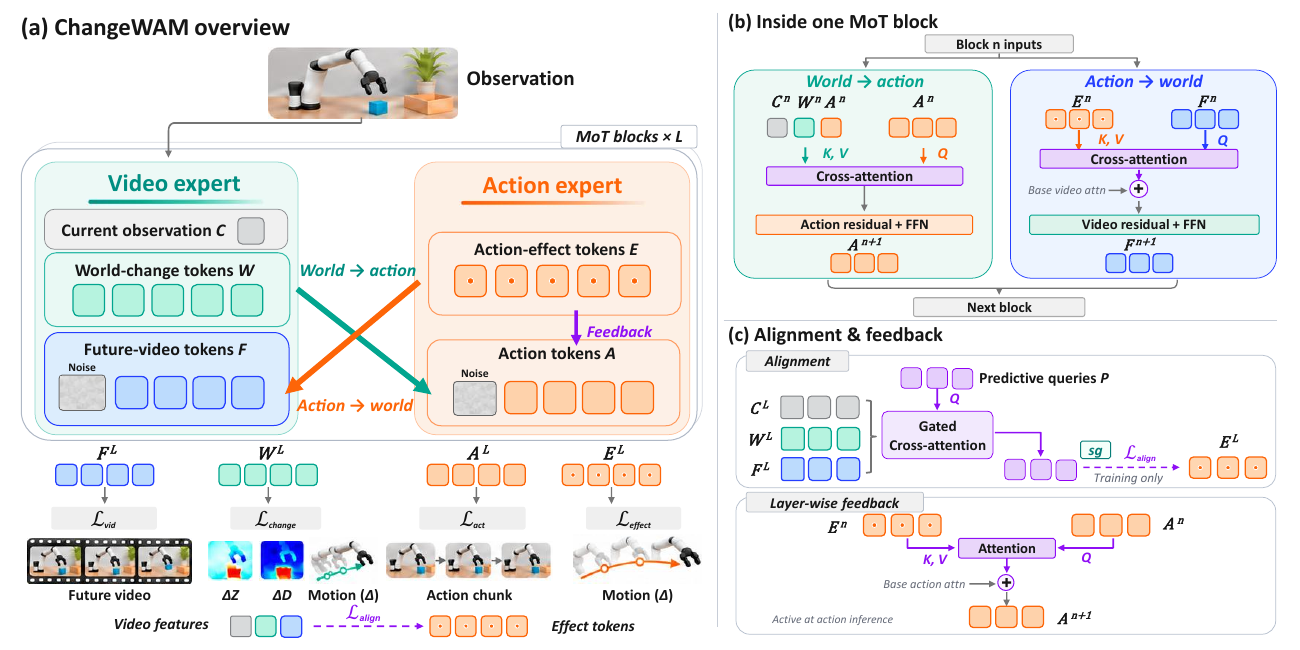}
    \vspace{-6mm}
    \caption{\textbf{\methodname architecture and learning objectives.}
    (a) We augment the coupled video and action experts while retaining their video and action denoising objective, $\mathcal L_{\mathrm{vid}}$ and $\mathcal L_{\mathrm{act}}$.
    World-change tokens $\mathbf W$ predict video, depth, and robot-motion changes under $\mathcal L_{\mathrm{change}}$; action-effect tokens $\mathbf E$ predict robot motion under $\mathcal L_{\mathrm{effect}}$.
    (b) Within a MoT block, world-change features guide action attention, while action-effect features add guidance to future-video attention. 
    (c) During training, predictive queries $\mathbf P$ aggregate current observation, world-change, and future-video features through gated cross-attention, providing a stop-gradient target for action-effect alignment. Layer-wise feedback allows action generation to use the aligned effect features without generating future video at inference.
    }
    \label{fig:overview}
     \vspace{-1mm}
\end{figure*}

% \vspace{-2mm}
\section{Related Work}
\label{sec:related}

\subsection{WAMs for Robotic Manipulation}
Two prominent paradigms for robotic manipulation are vision-language-action (VLA) models~\cite{openvla,pi0,pi05,qwenrobotmanip,wang25vqvla} and world-action models (WAMs)~\cite{uwm,lingbotva,fastwam}.
WAMs connect visual prediction with action generation through different forms of coupling~\cite{uwm,lingbotva,fastwam}.
One line of work constructs visual plans or predictive features to condition inverse dynamics models or action policies~\cite{unipi,vpp,vpp2,deltawam}.
Recent WAMs increasingly integrate visual dynamics and action generation through coupled diffusion or autoregressive modeling~\cite{uva,uwm,worldvla,videovla,dreamzero}.
This joint modeling can retain distinct video and action experts linked by attention~\cite{motus,motubrain,lingbotva,dit4dit}, or use a shared generative backbone that represents actions alongside visual tokens~\cite{cosmospolicy,patchwam}.
We focus on the former, where a central design question is how to organize the interaction between world and action experts.
Fast-WAM co-trains video and action generation while restricting the action expert's visual access to current-observation features~\cite{fastwam}.
Other designs condition action generation on predictive features or future-informed context~\cite{rift,internw0delta,placa,vjepapolicy}.
\methodname organizes expert interaction around change prediction in both directions: predicted world changes guide action generation, while motion inferred from action hypotheses guides future-video prediction.

% \vspace{-2mm}
\subsection{Additional Visual Priors for WAMs}
A complementary line of research enriches action learning with geometric, motion, and semantic information beyond RGB appearance~\cite{dreamwam,4dwam,dcwam}.
Within WAMs, such information can serve as additional prediction targets~\cite{dreamwam}, representation supervision~\cite{4dwam,internw0delta,agra,rgf}, or dedicated spatial experts~\cite{wsa1}.
DreamWAM combines RGB--motion latent denoising with geometry and semantic prediction, while 4D-WAM distills trajectory-field knowledge through motion and destination alignment~\cite{dreamwam,4dwam}.
InternW0-$\Delta$ combines semantic guidance from a frozen VLM with training-time distillation of 4D geometric and motion priors~\cite{internw0delta}.
WSA$_1$ introduces a dedicated 3D spatial expert to connect action-conditioned prediction of future 3D representations with 3D-guided action generation~\cite{wsa1}.
Related VLA methods also incorporate dynamic geometry: LaMP conditions action generation on a 3D scene-flow motion expert, while Track4Action distills world-centric 3D tracking features into policy-side queries~\cite{lamp,track4action}.
\methodname focuses on manipulation-related world changes, organizing bidirectional interaction within coupled video and action experts around explicitly supervised changes in appearance, depth, and robot motion.

\section{Method}
\label{sec:method}

We build \methodname on the Fast-WAM architecture~\cite{fastwam} (\Cref{fig:overview}), retaining its video and action denoising objectives (\Cref{sec:preliminaries}).
We introduce world-change tokens in the video expert, supervised to predict world changes and read by the action expert to guide action generation (\Cref{sec:world_to_action}).
In the action expert, we introduce action-effect tokens, supervised to predict end-effector motion and read by the video expert to guide future-video prediction (\Cref{sec:action_to_world}).
We further align the action-effect tokens with video-side predictive features and feed them back into the action expert to improve action generation (\Cref{sec:alignment_feedback}).
Our design can be incorporated into any Fast-WAM-style variants with coupled video and action experts, providing change-based bidirectional guidance for their existing prediction objectives.

\subsection{Preliminaries: Fast-WAM}
\label{sec:preliminaries}

Fast-WAM takes the current visual observation, robot state, and language instruction as input and jointly learns future-video prediction and action generation.
Its action output is an action chunk, a sequence of robot commands over a short horizon.
Its video expert and action expert are separately parameterized diffusion transformers, coupled through $L$ paired mixture-of-transformers (MoT) blocks.
A video variational autoencoder (VAE) encodes the observed and future frames into latent representations.
During training, the current-observation latents are kept clean and embedded as tokens $\mathbf C$, while future-video latents and action sequences are independently corrupted with noise and embedded as future-video tokens $\mathbf F$ and action tokens $\mathbf A$, respectively.

The video expert processes $\mathbf C$ and $\mathbf F$ together: current tokens attend to one another, and future tokens read both current and future tokens.
In the action expert, action tokens $\mathbf A$ attend to one another and to current-observation tokens $\mathbf C$ at each MoT block, giving action generation access to the observed scene.
The language instruction is encoded into text embeddings, and the robot state is projected into a conditioning token.
These are concatenated into a shared conditioning sequence that both experts read through their own cross-attention layers.

At the final block, prediction heads map $\mathbf F^L$ and $\mathbf A^L$ to video and action denoising velocities, respectively.
They are trained with flow-matching losses $\mathcal L_{\mathrm{vid}}$ and $\mathcal L_{\mathrm{act}}$, using the clean future-video latents and recorded action sequences to construct the denoising targets.
The resulting video latents can be decoded into future RGB frames, and the denoised action sequence provides the robot commands.
At inference, the action expert can generate an action chunk by iteratively denoising action tokens from Gaussian noise while reading the current-observation features, without generating future video.

Superscripts $n$ and $L$ denote features at MoT block $n$ and after the final block, respectively.
In the following equations, $\operatorname{Attn}(\text{query},\text{key},\text{value})$ denotes attention and $[\,;\,]$ denotes token concatenation.
We use $\widetilde{\mathbf X}^{n}$ for the attention output of token group $\mathbf X^n$; the expert's output projection, residual updates, conditioning cross-attention, and feed-forward layers then produce $\mathbf X^{n+1}$.

\subsection{World Changes for Action Generation}
\label{sec:world_to_action}

Future RGB prediction provides dense supervision, but manipulation depends particularly on what moves, how geometry changes, and how these changes relate to robot behavior.
We therefore introduce a spatial grid of learnable world-change tokens $\mathbf W$ in the video expert, alongside the current-observation tokens $\mathbf C$ and future-video tokens $\mathbf F$.
At each MoT block, current and future-video tokens retain their original attention:
\begin{equation}
\begin{aligned}
    \widetilde{\mathbf C}^{n}&=\operatorname{Attn}\!\left(
        \mathbf C^n,\mathbf C^n,\mathbf C^n\right),\\
    \widetilde{\mathbf F}^{n}&=\operatorname{Attn}\!\left(
        \mathbf F^n,[\mathbf C^n;\mathbf F^n],[\mathbf C^n;\mathbf F^n]\right).
\end{aligned}
    \label{eq:video_attention}
\end{equation}

Within the same video block, world-change tokens attend to current visual tokens and to one another:
\begin{equation}
    \widetilde{\mathbf W}^{n}=\operatorname{Attn}\!\left(
        \mathbf W^n,[\mathbf C^n;\mathbf W^n],[\mathbf C^n;\mathbf W^n]\right).
    \label{eq:world_change_attention}
\end{equation}
% Thus, $\mathbf W$ learns predictive features from the current observation alongside future-video denoising.

At the final layer, $\mathbf F^L$ retains the original video flow-matching supervision:
\begin{equation}
    \mathcal L_{\mathrm{vid}}=
    \ell_{\mathrm{FM}}\!\left(h_{\mathrm{vid}}(\mathbf F^L),\mathbf v_{\mathrm{vid}}\right).
    \label{eq:video_loss}
\end{equation}
Here, $h_{\mathrm{vid}}$ maps the final future-video tokens to a denoising velocity, and $\mathbf v_{\mathrm{vid}}$ is the target velocity.
$\ell_{\mathrm{FM}}$ denotes the mean-squared velocity prediction error.

Alongside this objective, three prediction heads $h_z$, $h_d$, and $h_m$ use $\mathbf W^L$ to estimate video differences, depth differences, and robot-motion changes across the prediction horizon.

Let $\mathbf Z_j$ denote clean video latents and $D_k$ the depth maps supplied by a frozen geometric teacher, indexed on their respective temporal grids.
We construct difference targets as
\begin{equation}
    \Delta\mathbf Z_j=\mathbf Z_j-\mathbf Z_{j-1},
    \qquad
    \Delta D_k=\frac{D_k-D_{k-1}}{\operatorname{med}_{\mathrm{valid}}(D_0)}.
    \label{eq:change_targets}
\end{equation}
Robot motion is described by $\mathbf m_i=[\Delta\mathbf p_i;\Delta\mathbf r_i]$, collecting end-effector translation and rotation changes at future offset $\delta_i$.
Writing $\mathbf m$ for the motion targets across future offsets, we group video-side change supervision as
\begin{equation}
\begin{aligned}
    \mathcal L_{\mathrm{change}}={}&
    \lambda_z\,\ell_{2}(h_z(\mathbf W^L),\Delta\mathbf Z)\\
    &+\lambda_d\,\ell_{\mathrm{Huber}}(h_d(\mathbf W^L),\Delta D)\\
    &+\lambda_{m}\,\ell_{\mathrm{Huber}}
    (h_{m}(\mathbf W^L), \mathbf m),
\end{aligned}
    \label{eq:change_loss}
\end{equation}
where $\ell_2$ denotes squared error and $\ell_{\mathrm{Huber}}$ denotes Huber loss.
The $\lambda$ coefficients balance the three prediction tasks; their values appear in the supplementary material~\Cref{sec:appendix_losses}.

Together, these tasks teach $\mathbf W$ what changes in the scene and where manipulation-related motion occurs.
To use these predictive features for action generation, we extend the keys and values of the existing action attention with $\mathbf W$ (\Cref{fig:overview}(b)):
\begin{equation}
    \widetilde{\mathbf A}^{n}=\operatorname{Attn}\!\left(
        \mathbf A^n,[\mathbf C^n;\mathbf W^n;\mathbf A^n],
        [\mathbf C^n;\mathbf W^n;\mathbf A^n]\right).
    \label{eq:world_to_action}
\end{equation}
Thus, action denoising uses both the observed scene and change-related features to generate actions consistent with the anticipated evolution of the scene.

\subsection{Action Effects for World Prediction}
\label{sec:action_to_world}

The reverse direction asks how the world should evolve under the action currently being considered.
We therefore append learnable action-effect tokens $\mathbf E$ to the action stream to infer the motion implied by the action  hypothesis.
Within each action block, they read the action hypothesis, current visual context, and one another:
\begin{equation}
    \widetilde{\mathbf E}^{n}=\operatorname{Attn}\!\left(
        \mathbf E^n,[\mathbf C^n;\mathbf A^n;\mathbf E^n],
        [\mathbf C^n;\mathbf A^n;\mathbf E^n]\right).
    \label{eq:effect_readout}
\end{equation}

At the final layer, $\mathbf A^L$ retains the original action flow-matching supervision:
\begin{equation}
    \mathcal L_{\mathrm{act}}=
    \ell_{\mathrm{FM}}\!\left(h_{\mathrm{act}}(\mathbf A^L),\mathbf v_{\mathrm{act}}\right).
    \label{eq:action_loss}
\end{equation}
Here, $h_{\mathrm{act}}$ maps the final action tokens to a denoising velocity, and $\mathbf v_{\mathrm{act}}$ is the  target velocity.
  
Alongside action denoising, a motion prediction head $h_m$ uses $\mathbf E^L$ to estimate the end-effector pose changes $\mathbf m$:
\begin{equation}
    \mathcal L_{\mathrm{effect}}=\lambda_m\,
    \ell_{\mathrm{Huber}}(h_m(\mathbf E^L),\mathbf m).
    \label{eq:effect_loss}
\end{equation}
This objective gives $\mathbf E$ an explicit physical meaning: the robot motion implied by the current action hypothesis.
To use these features for world prediction, future-video tokens query the action-effect tokens (\Cref{fig:overview}(b)), adding motion guidance to the video attention in Eq.~\ref{eq:video_attention}:
\begin{equation}
    \widetilde{\mathbf F}^{n}\leftarrow\widetilde{\mathbf F}^{n}+
    \operatorname{Attn}\!\left(\mathbf F^n,\mathbf E^n,\mathbf E^n\right).
    \label{eq:action_to_world}
\end{equation}
Thus, future-video prediction uses action-derived motion features to anticipate how the scene evolves under the action hypothesis.

\subsection{Predictive Feature Alignment and Feedback}
\label{sec:alignment_feedback}

Action-guided world prediction improves the video expert's predictive representations.
To further improve action generation with these representations, we align the action-effect tokens with video-side predictive features and feed the aligned tokens back into the action expert (\Cref{fig:overview}(c)).
In this way, action generation can benefit from the video expert's predictive knowledge without generating future video at inference.

We introduce learnable predictive tokens $\mathbf P$ to extract information from the video expert through gated cross-attention:
\begin{equation}
    \mathbf P\leftarrow\mathbf P+
    \sum_{X\in\{C,W,F\}}
    \alpha_X\,
    \operatorname{Attn}\!\left(
        \mathbf P,\mathbf X^L,\mathbf X^L\right),
    \label{eq:predictive_readout}
\end{equation}
where $\boldsymbol{\alpha}_X$ denotes learnable gates for each feature source.
The readout combines current visual context, world-change features, and future-video features into predictive summaries.
The predictive queries are supervised to predict robot motion using the same prediction head $h_{m}$ as the action-effect tokens. 
Implementation details are provided in the supplementary material~\Cref{sec:appendix_architecture_details}.

We align the action-effect tokens with these predictive features using a stop-gradient target:
\begin{equation}
    \mathcal L_{\mathrm{align}}=
    1-\cos\!\left(\mathbf E^L,\operatorname{sg}(\mathbf P)\right),
    \label{eq:predictive_alignment}
\end{equation}
where $\operatorname{sg}$ denotes stop-gradient.
Alignment is computed in a shared feature space, with cosine similarity averaged over valid token pairs.

Finally, the action expert reads the aligned effect tokens through cross-attention:
\begin{equation}
    \widetilde{\mathbf A}^{n}\leftarrow\widetilde{\mathbf A}^{n}+
    \operatorname{Attn}\!\left(\mathbf A^n,\mathbf E^n,\mathbf E^n\right).
    \label{eq:effect_feedback}
\end{equation}
This feedback allows visual predictive knowledge to inform action denoising.
The predictive readout is used only during training; action inference uses the learned effect tokens and feedback pathway without generating future video.

The original video and action objectives together form
\begin{equation}
    \mathcal L_{\mathrm{WAM}}=\mathcal L_{\mathrm{vid}}+\mathcal L_{\mathrm{act}}.
    \label{eq:wam_loss}
\end{equation}
Our main learning objective combines this WAM supervision, video-side change prediction, action-side effect prediction, and feature alignment:
\begin{equation}
\begin{aligned}
    \mathcal L_{\mathrm{main}}={}&\mathcal L_{\mathrm{WAM}}+
    \mathcal L_{\mathrm{change}}+\mathcal L_{\mathrm{effect}}
    +\lambda_{\mathrm{align}}\mathcal L_{\mathrm{align}}.
\end{aligned}
    \label{eq:total_objective}
\end{equation}

At inference, the video expert processes the current observation to obtain visual and world-change features, whose layer-wise keys and values are cached.
The action expert denoises an action chunk using cached visual features, while deriving action-effect features from its current action hypothesis at each  denoising step.
The feedback pathway remains active, allowing action generation to use the predictive information learned through alignment.
Auxiliary prediction heads are unused during action inference, and future-video generation is not required.

\section{Experiments}
\label{sec:experiments}
We evaluate \methodname through out-of-distribution manipulation benchmarks (\Cref{sec:ood_results}), comparisons with related methods that improve manipulation learning through auxiliary supervision (\Cref{sec:comparison_methods}), and ablations of the learning objectives  (\Cref{sec:design_analysis}).
We further examine how action guidance improves video prediction (\Cref{sec:world_prediction_results}) and assess inference efficiency (\Cref{sec:efficiency}).
Finally, we evaluate \methodname on real-world manipulation tasks under varying scene conditions (\Cref{sec:real_robot}).

\subsection{Experimental Setup}
\label{sec:experimental_setup}

\paragraph{Benchmarks.}
We train on LIBERO demonstrations~\cite{libero} and evaluate on LIBERO-Plus and LIBERO-Pro without additional fine-tuning.
LIBERO-Plus~\cite{liberoplus} covers changes in object layout, camera viewpoint, robot initialization, language, lighting, background, and sensor noise.
We evaluate all 10,030 cases across Spatial, Object, Goal, and Long.
Our LIBERO-Pro~\cite{liberopro} evaluation covers Object, Position, Semantic, and Task shifts, totaling 8,000 episodes across 160 task variants.
For RoboTwin 2.0~\cite{robotwin2}, we train on 2,500 clean demonstrations from 50 bimanual tasks and evaluate 100 valid episodes per task under randomized scenes and unseen instructions.

\paragraph{Baselines.}
\label{sec:baselines}
We organize the comparisons into two groups.
\emph{Add-on evaluations} examine the gains from applying the mechanism of \methodname to Fast-WAM~\cite{fastwam}, 4D-WAM~\cite{4dwam}, and InternW0-$\Delta$~\cite{internw0delta} (\Cref{sec:ood_results}).
For the 4D-WAM add-on, \methodname only incorporates its destination and motion trace-alignment losses, while retaining our expert-interaction design and action inference without future-video denoising.
\emph{Related-method comparisons} assess \methodname against other WAM enhancement mechanisms that improve manipulation learning through auxiliary supervision: 4D-WAM, Track4Action, DreamWAM, and WSA$_1$-L~\cite{4dwam,track4action,dreamwam,wsa1} (\Cref{sec:comparison_methods}).
We implement their corresponding mechanisms to Fast-WAM~\cite{fastwam}.
WSA$_1$-L is evaluated using its official checkpoints and training budget.

\paragraph{Implementation details.}
\label{sec:implementation_details}
\emph{Inputs and supervision.}
LIBERO uses a $224\times448$ two-camera canvas and seven-dimensional actions; RoboTwin 2.0 uses $384\times320$ three-camera input and 14-dimensional joint/gripper actions.
A frozen Pi3X teacher~\cite{pi3} generates depth targets offline.
Robot-motion targets are derived from recorded end-effector states in LIBERO and from joint commands using the ALOHA–AgileX kinematic model in RoboTwin 2.0.
\emph{Training.}
To examine gains from different starting points, we consider two initialization settings while keeping the expert architecture unchanged.
The standard setting uses Wan video weights and interpolated ActionDiT weights; the improved setting transfers compatible pretrained parameters from InternW0-$\Delta$~\cite{internw0delta}.
Controlled comparisons use the same training data, initialization, and downstream training budget within each setting.
LIBERO models are trained for 25,000 updates with global batch 128; RoboTwin 2.0 models use 21,480 updates with global batch 256.
Both use learning rate $10^{-4}$, cosine decay, weight decay $0.01$, BF16, and training seed 42.
During training on LIBERO and RoboTwin 2.0 clean demonstrations for these controlled comparisons, we do not use data augmentation designed to mimic test-time distribution shifts.
\emph{Inference.}
Policies in the controlled comparisons predict 32-step action chunks, execute 10 actions before replanning, and use 10 denoising steps.
\methodname uses cached action-only inference on LIBERO-Plus, LIBERO-Pro, and RoboTwin 2.0, without denoising future-video latents.
Further architecture, objective weights, and evaluation details appear in the supplementary material.

\begin{table}[t]
\centering
\small
\setlength{\tabcolsep}{2.5pt}
\caption{\textbf{ChangeWAM as an add-on to existing world-action models.} $^\dagger$ denotes improved initialization.}
\vspace{-2mm}
\label{tab:ood_results}
\begin{tabular}{@{}llrr@{}}
\toprule
Base WAM & Variant & LIBERO-Plus & LIBERO-Pro \\
\midrule
\multirow{2}{*}{Fast-WAM~\cite{fastwam}} & Base & 49.66 & 42.71 \\
 & \textbf{+Ours} & \textbf{79.02} \scriptsize (+29.36) & \textbf{51.06} \scriptsize (+8.35) \\
\midrule
\multirow{2}{*}{4D-WAM~\cite{4dwam}} & Base & 69.00 & 47.76 \\
 & \textbf{+Ours} & \textbf{80.13} {\scriptsize (+11.13)} & \textbf{51.69} {\scriptsize (+3.93)} \\
\midrule
\multirow{2}{*}{4D-WAM$^\dagger$} & Base & 80.37 & 56.88 \\
 & \textbf{+Ours} & \textbf{84.68} {\scriptsize (+4.32)} & \textbf{60.01} {\scriptsize (+3.14)} \\
\midrule
\multirow{2}{*}{InternW0-$\Delta$~\cite{internw0delta}} & Base & 92.78 & 64.88 \\
 & \textbf{+Ours} & \textbf{93.93} {\scriptsize (+1.15)} & \textbf{65.69} {\scriptsize ($+0.81$)} \\
\bottomrule
\end{tabular}
\vspace{-1mm}
\end{table}

\begin{table}[t]
\centering
\small
\setlength{\tabcolsep}{3pt}
\caption{\textbf{LIBERO-Plus perturbation breakdown for the 4D-WAM add-on.} Average success rates (\%).}
% \vspace{-2mm}
\label{tab:plus_perturbations}
\begin{tabular}{lrrr}
\toprule
Perturbation & 4D-WAM & +Ours & Gain (pp) \\
\midrule
\multicolumn{4}{l}{\emph{Standard initialization}} \\
Background texture & 59.29 & \textbf{75.65} & +16.36 \\
Camera viewpoint & 42.96 & \textbf{64.29} & +21.33 \\
Language & \textbf{92.32} & 92.26 & -0.07 \\
Lighting & 92.73 & \textbf{95.97} & +3.24 \\
Object layout & 79.34 & \textbf{83.54} & +4.20 \\
Robot initial state & 62.19 & \textbf{68.90} & +6.71 \\
Sensor noise & 57.71 & \textbf{82.57} & +24.86 \\
\midrule
\multicolumn{4}{l}{\emph{Improved initialization}} \\
Background texture & 86.99 & \textbf{91.91} & +4.93 \\
Camera viewpoint & 50.84 & \textbf{60.85} & +10.01 \\
Language & \textbf{97.01} & 96.94 & -0.07 \\
Lighting & 98.25 & \textbf{98.77} & +0.53 \\
Object layout & 85.57 & \textbf{87.48} & +1.90 \\
Robot initial state & 89.10 & \textbf{90.39} & +1.29 \\
Sensor noise & 62.52 & \textbf{72.77} & +10.24 \\
\bottomrule
\end{tabular}

\vspace{-1mm}
\end{table}

\begin{table}[t]
\centering
\small
\setlength{\tabcolsep}{5pt}
\caption{\textbf{ChangeWAM as an add-on on RoboTwin 2.0.} Base denotes the Fast-WAM-based implementation of 4D-WAM. $^\dagger$ denotes improved initialization.}
\vspace{-2mm}
\label{tab:robotwin_results}
\begin{tabular}{@{}llrr@{}}
\toprule
Base WAM & Variant & C2C & C2R \\
\midrule
\multirow{2}{*}{4D-WAM$^\dagger$} & Base & 82.04 & 22.46 \\
 & \textbf{+Ours} & \textbf{83.18} {\scriptsize (+1.14)} & \textbf{41.22} {\scriptsize (+18.76)} \\
\bottomrule
\end{tabular}
\vspace{-1mm}
\end{table}

\begin{table}[t]
\centering
\small
\setlength{\tabcolsep}{4pt}
\caption{\textbf{Comparison with related methods on LIBERO-Plus.} Average success rates (\%). $^*$ denotes released  references with their own training budgets. Fast-WAM provides a shared reference.}
\vspace{-2mm}
\label{tab:plus_comparison}
\begin{tabular}{lrrrrr}
\toprule
Method & Spatial & Object & Goal & Long & Avg. \\
\midrule
Fast-WAM~\cite{fastwam} & 52.21 & 69.06 & 37.32 &40.06 & 49.66\\
\midrule
4D-WAM~\cite{4dwam} & 79.43 & 74.70 & 58.05 & 63.83 & 69.00 \\
Track4Action~\cite{track4action} & 80.22 & 72.95 & 57.31 & 67.69 & 69.54 \\
DreamWAM~\cite{dreamwam} & 84.80 & 77.76 & 61.79 & 65.30 & 72.41 \\
WSA$_1$-L$^*$~\cite{wsa1} & 83.85 & 80.66 & 60.09 & 72.73 & 74.33 \\
\textbf{\methodname} & \textbf{87.34} & \textbf{89.41} & \textbf{66.22} & \textbf{73.12} & \textbf{79.02} \\
\bottomrule
\end{tabular}
\vspace{-1mm}
\end{table}

\begin{table*}[t]
\centering
\small
\setlength{\tabcolsep}{7pt}
\caption{\textbf{Progressive design analysis on LIBERO-Plus.} The grouped columns expand $\mathcal L_{\mathrm{change}}$. Rows compare complete configurations with their associated auxiliary supervision.}
\vspace{-2mm}
\label{tab:design_comparison}
\begin{tabular}{lcccccr}
\toprule
\multirow{2}{*}{Training configuration} & \multicolumn{3}{c}{$\mathcal L_{\mathrm{change}}$ (Eq.~\ref{eq:change_loss})} & \multirow{2}{*}{$\mathcal L_{\mathrm{effect}}$} & \multirow{2}{*}{$\mathcal L_{\mathrm{align}}$} & \multirow{2}{*}{Success (\%)} \\
\cmidrule(lr){2-4}
& Video & Depth & Robot motion & & & \\
\midrule
Baseline & -- & -- & -- & -- & -- & 69.00 \\
+ Video change supervision & $\checkmark$ & -- & -- & -- & -- & 73.83 \\
+ Depth change supervision & $\checkmark$ & $\checkmark$ & -- & -- & -- & 75.50 \\
+ Robot-motion supervision & $\checkmark$ & $\checkmark$ & $\checkmark$ & -- & -- & 77.28 \\
+ $\mathcal L_{\mathrm{effect}}$ and reverse injection & $\checkmark$ & $\checkmark$ & $\checkmark$ & $\checkmark$ & -- & 78.70 \\
\textbf{+ $\mathcal L_{\mathrm{align}}$ and feedback (\methodname)} & $\checkmark$ & $\checkmark$ & $\checkmark$ & $\checkmark$ & $\checkmark$ & \textbf{80.13}\\
\midrule
\multicolumn{6}{l}{\methodname with absolute future targets} & \textit{76.57} \\
\bottomrule
\end{tabular}
\vspace{-2mm}
\end{table*}

\subsection{Out-of-Distribution Generalization}
\label{sec:ood_results}

\paragraph{OOD results on LIBERO-Plus and LIBERO-Pro.}
\Cref{tab:ood_results} compares the base models introduced in \Cref{sec:baselines} with their \methodname add-ons.
For Fast-WAM, applying the modules proposed in \methodname improves success by 29.36 and 8.35 percentage points on LIBERO-Plus and LIBERO-Pro, respectively.
For 4D-WAM, we examine both standard and improved initialization.
With standard initialization, applying \methodname mechanism raises success from 69.00\% to 80.13\% on LIBERO-Plus and from 47.76\% to 51.69\% on LIBERO-Pro, gains of 11.13 and 3.93 percentage points, respectively.
The gains persist with improved initialization: success increases from 80.37\% to 84.68\% on LIBERO-Plus and from 56.88\% to 60.01\% on LIBERO-Pro.
Applied to InternW0-$\Delta$, our interaction adaptation further improves success from 92.78\% to 93.93\% on LIBERO-Plus and from 64.88\% to 65.69\% on LIBERO-Pro.
These gains across base WAMs and initialization settings suggest that change-based expert interaction can complement existing spatial supervision and pretrained knowledge to improve out-of-distribution manipulation.

The perturbation breakdown in \Cref{tab:plus_perturbations} compares the 4D-WAM baseline with 4D-WAM + \methodname on LIBERO-Plus, under both initialization settings.
With standard initialization, the largest improvements occur under sensor noise (+24.86 percentage points), camera viewpoint changes (+21.33), and background texture changes (+16.36).
With improved initialization, the gains under sensor noise and camera changes remain substantial at +10.24 and +10.01 points, respectively.
These results highlight improved robustness to visual changes that alter the observations available for control.

\paragraph{OOD results on RoboTwin 2.0.}
We further evaluate the add-on results using the 4D-WAM configuration with improved initialization.
After training on clean demonstrations, adding \methodname increases success under randomized scenes and unseen instructions from 22.46\% to 41.22\%, a gain of 18.76 percentage points (\Cref{tab:robotwin_results}).
Evaluating the same checkpoints on clean scenes also increases success from 82.04\% to 83.18\%.
The larger gain under randomization highlights the benefit of change-based guidance for generalization to unseen scene conditions.

\subsection{Comparison with Auxiliary World Supervision}
\label{sec:comparison_methods}

We also compare different mechanisms that can be incorporated into a WAM to improve manipulation learning.
\Cref{tab:plus_comparison} reports results across the four suites.
Under the common training setup, 4D-WAM, Track4Action, and DreamWAM obtain average success rates of 69.00\%, 69.54\%, and 72.41\%, respectively; \methodname reaches 79.02\% and improves all four suites over these comparisons.
The released WSA$_1$-L policy obtains 74.33\% on the same cases.
% The table also lists the published Fast-WAM-Joint overall score.
This comparison shows the effectiveness of organizing auxiliary supervision around the changes exchanged between the video and action experts.

\begin{figure}[t]
    \centering
    \includegraphics[width=\linewidth]{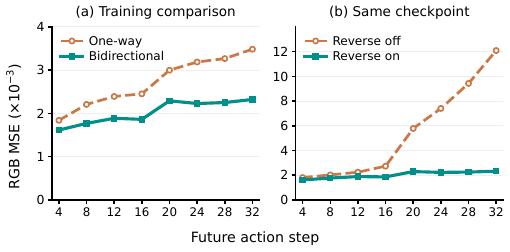}
     \vspace{-6mm}
   \caption{\textbf{Action guidance reduces future-video prediction error.}
    RGB MSE on LIBERO demonstration clips (lower is better).
    (a)~Separately trained one-way and bidirectional models.
    (b)~Reverse guidance off versus on in the same bidirectional checkpoint.}
    \label{fig:prediction_horizons}
    \vspace{-1mm}
\end{figure}

\begin{figure*}[t]
    \centering
    \includegraphics[width=\linewidth]{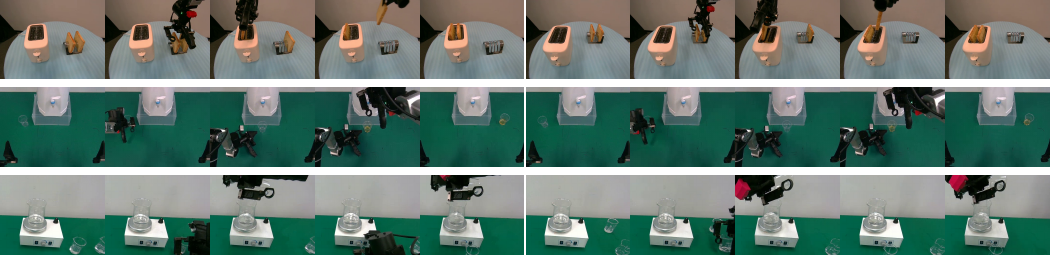}
    \vspace{-2mm}
    
    \caption{\textbf{Real-world manipulation under varied object placements.} Rows correspond to Toast, Water, and Liquid transfer, respectively, with two layouts per task and five chronological frames per rollout. 
    }
    \label{fig:real_robot_rollouts}
     \vspace{-1mm}
\end{figure*}

\subsection{Ablation of the Learning Objectives}
\label{sec:design_analysis}

We progressively extend 4D-WAM to examine world-change supervision, action-guided world prediction, and predictive feature alignment with feedback (\Cref{tab:design_comparison}).
All configurations use standard Wan initialization and the same training budget.

\vspace{-1mm}
\paragraph{World changes for action generation.}
Adding video-change supervision raises success from 69.00\% to 73.83\%.
Adding depth-change supervision further improves success to 75.50\%.
Adding robot-motion change supervision further increases success to 77.28\%.
These gains suggest that anticipating scene changes provides useful guidance for action generation, and that grounding these predictions in robot motion makes them more relevant to manipulation.

\vspace{-1mm}
\paragraph{Action effects for world prediction.}
The basic bidirectional variant supervises action-effect features with robot-motion changes and injects these features into the video expert through attention to guide future-video prediction.
This action-to-world guidance increases success to 78.70\%.
Its benefit to video prediction is examined directly in \Cref{sec:world_prediction_results}.

\vspace{-1mm}
\paragraph{Predictive feature alignment and feedback.}
The full design aligns action-effect features with video-side predictive features and feeds them back into action denoising, reaching 80.13\% success.
Overall, \methodname improves success by 11.13 percentage points over the base model 4D-WAM.
Together, these progressive comparisons support the effectiveness of the learning objectives and interaction designs introduced in \methodname.

\vspace{-1mm}
\paragraph{Change prediction effectiveness.}
To isolate the role of change-based supervision, we also compare predicting changes with directly predicting future quantities.
We replace auxiliary video and depth differences with future video latents and depth, and replace end-effector pose changes with absolute future poses in both prediction directions.
This reduces LIBERO-Plus success by \textit{3.56} percentage points, supporting the benefit of change-based supervision for action generalization.

\subsection{Action-Guided World Prediction}
\label{sec:world_prediction_results}

We evaluate future-video prediction on 200 LIBERO demonstration clips across 40 tasks (\Cref{fig:prediction_horizons}).
Bidirectional training reduces final-frame RGB MSE by \textbf{33.36\%} relative to one-way training.
Enabling reverse guidance in the same bidirectional checkpoint reduces final-frame error by \textbf{80.79\%}.
Both comparisons show lower error throughout the prediction horizon, supporting the role of reverse motion guidance in improving future-video prediction.

\subsection{Inference Efficiency}
\label{sec:efficiency}

\begin{table}[t]
\centering
\small
\setlength{\tabcolsep}{2.5pt}
\caption{\textbf{Inference cost of ChangeWAM add-ons.}
  }
  \vspace{-2mm}
\label{tab:inference_cost}
\begin{tabular}{@{}llcr@{}}
\toprule
Base WAM & Variant & \shortstack{Future-video\\denoising} & Latency (ms) \\
\midrule
\multirow{2}{*}{Fast-WAM~\cite{fastwam}} & Base & No & 528.2 \\
 & +\methodname & No & 639.6 \\
\midrule
\multirow{2}{*}{4D-WAM~\cite{4dwam}} & Base & Yes & 1019.8 \\
 & +\methodname & No & 639.6 \\
\bottomrule
\end{tabular}
\vspace{-1mm}
\end{table}

\begin{table}[t]
\centering
\small
\setlength{\tabcolsep}{3pt}
\caption{\textbf{Real-world manipulation on Acone.} $^\dagger$ denotes improved initialization.}
\vspace{-2mm}
\label{tab:real_robot}
\begin{tabular}{lccc}
\toprule
Method  & Toast & Liquid Transfer & Water \\
\midrule
4D-WAM  & 15\% & 20\% & -- \\
+\methodname  & 40\% & 40\% & -- \\
\midrule
4D-WAM$^\dagger$  & 45\% & 40\%  & 10\%\\
+\methodname  & 60\% & 70\% & 40\% \\
\bottomrule
\end{tabular}
\vspace{-1mm}
\end{table}

We evaluate the inference cost of applying \methodname mechanism to Fast-WAM and 4D-WAM under matched hardware and settings (\Cref{tab:inference_cost}).
On Fast-WAM, the add-on increases latency from 528.2 to 639.6\,ms per 32-action chunk, reflecting the extra computation for action-effect updates and feedback attention.
On 4D-WAM, it reduces latency from 1,019.8 to 639.6\,ms, a reduction of 37.3\%, because action generation no longer requires jointly denoising future video and attending to its evolving features.
Predictive feature alignment and feedback allow \methodname to use video-side predictive knowledge for action generation without denoising future video at inference.

\subsection{Real-World Manipulation}
\label{sec:real_robot}

We study real-world manipulation on the Acone bimanual robot with three tasks: Toast, Liquid transfer, and Water.
Each task uses a separately trained policy. 
We apply \methodname mechanism to 4D-WAM and compare the base and add-on policies.
Toast involves placing bread into a toaster under three settings: familiar bread-rack placements with the toaster fixed (\emph{easy}), unseen toaster positions with the bread rack easy to grasp (\emph{medium}), and unseen positions of both objects (\emph{hard}).
Liquid transfer uses beaker placements within (\emph{easy}) or outside (\emph{hard}) the training range.
Water requires grasping a cup, positioning it beneath a dispenser, and filling it to a specified level mark. 
We compare the base and add-on policies under both initializations for Toast and Liquid transfer; given its greater difficulty, Water is evaluated only with the improved pretrained initialization.
Each evaluated policy receives 20 trials per task: 6/7/7 easy/medium/hard trials for Toast, 10/10 easy/hard trials for Liquid transfer, and 20 trials in the single Water setting.
Corresponding base and add-on trials use the same scene layouts (\Cref{tab:real_robot}); \Cref{fig:real_robot_rollouts} shows selected execution sequences.

The add-on improves success in every evaluated task--initialization pair (\Cref{tab:real_robot}).
With improved initialization, the gains are 15, 30, and 30 percentage points on Toast, Liquid transfer, and Water, respectively.

\section{Conclusion}
\label{sec:conclusion}

We presented \methodname, a framework that organizes interaction between world and action experts through change-based bidirectional guidance.
Predicted world changes inform action generation, while motion inferred from action candidates guides future-video prediction.
Visual, geometric, and robot-motion supervision ground these predictions in changes relevant to manipulation.
Predictive feature alignment and feedback further allow action generation to benefit from better video-side predictive knowledge without requiring future-video generation at inference.
Experiments across multiple base WAMs and three out-of-distribution benchmarks demonstrate improved manipulation performance.
These results highlight change prediction as a shared basis for connecting how the world may evolve with how a robot should act, enabling world prediction and action generation to support one another in learning generalizable robot behavior.

\label{page:lastmain}
% \clearpage
\section*{LLM Usage Statement}
We used OpenAI ChatGPT and Codex to assist with manuscript drafting and editing, and code development and debugging.
We also discussed the methodology and experimental design with Codex and incorporated its feedback into refinements, while the overall research process remained human-led.
OpenAI image-generation tools were used to create illustrative assets for conceptual figures.
The authors take full responsibility for the scientific claims, experimental results, citations, and final content.
{\small
\bibliographystyle{ieeenat_fullname}
\bibliography{main}
}
\ifwithappendix
\clearpage
\appendix
\maketitlesupplementary

This supplementary material provides further details and analyses of \methodname.
\Cref{sec:appendix_architecture_details} describes the computation within a MoT block and the training of the predictive queries $\mathbf P$.
\Cref{sec:appendix_impl} specifies target construction, loss weights, and related-method implementations.
\Cref{sec:appendix_analysis} presents additional policy results and mechanism analyses.

\section{Implementation Details}
\label{sec:appendix_architecture_details}

\subsection{Computation Within a MoT Block}
\label{sec:appendix_architecture}

\begin{figure*}[t]
    \centering
    \includegraphics[width=\textwidth]{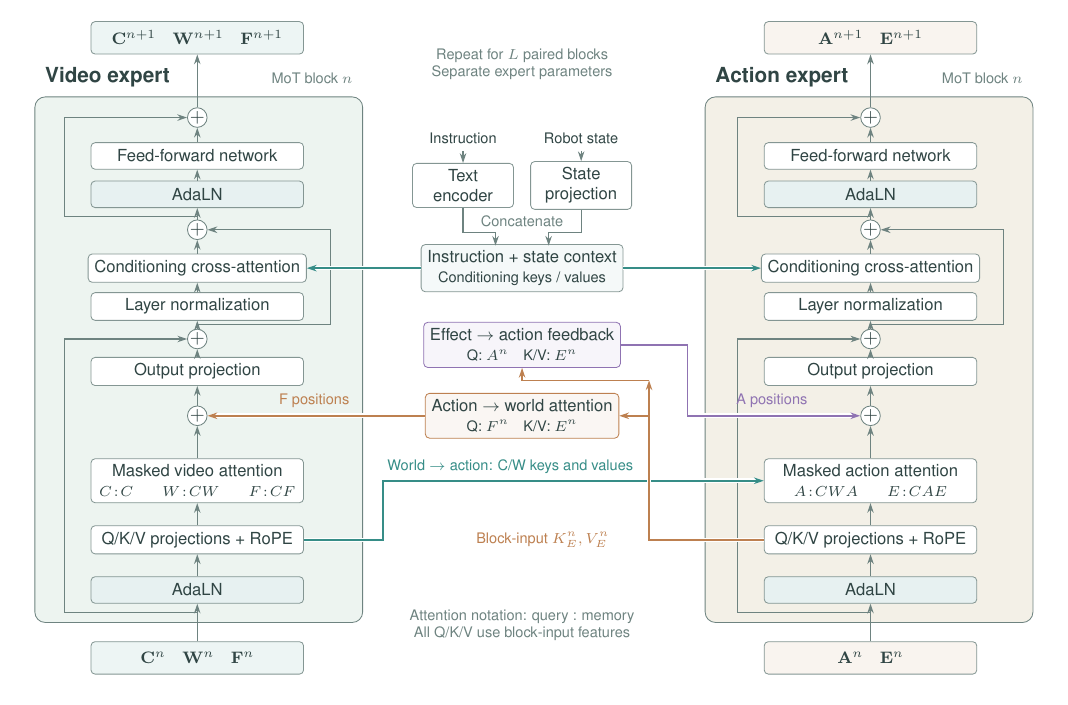}
    \caption{\textbf{Attention flow within a MoT block.}
    Read each expert from bottom to top; side paths indicate residual connections.
    All attention branches use queries, keys, and values projected from the same block inputs.
    Current-observation and world-change features enter the action attention memory, while action-effect features supply reverse guidance to future-video tokens and feedback to action tokens.
    The outputs are combined before each expert's own output projection, conditioning cross-attention, and feed-forward network.
    Instruction and robot-state embeddings form the conditioning memory.
    }
    \label{fig:supp_mot_flow}
\end{figure*}

\paragraph{Token streams and projections.}
Each MoT block processes two streams with separate parameters: the video expert takes current-observation tokens $\mathbf C^n$, world-change tokens $\mathbf W^n$, and future-video tokens $\mathbf F^n$; the action expert takes action tokens $\mathbf A^n$ and action-effect tokens $\mathbf E^n$.
The token groups within each stream are processed together, with attention masks specifying which features each group can read.
Across our Fast-WAM-based LIBERO and RoboTwin configurations, the paired expert blocks repeat $L=30$ times, with 32 action tokens and eight effect tokens associated with offsets $4,8,\ldots,32$.
The visual token counts depend on the input resolution: the $224\times448$ LIBERO canvas yields 98 current-observation tokens, 98 world-change tokens, and 196 future-video tokens, while the $384\times320$ RoboTwin canvas yields 120, 120, and 240 tokens, respectively.

At the start of block $n$, each expert applies adaptive layer normalization and projects its tokens into queries, keys, and values, followed by the backbone's query/key normalization and rotary position embeddings.
The experts project into a common attention space, allowing one expert's queries to read the other's keys and values.
We write token features directly as the arguments of $\operatorname{Attn}$, with their query, key, and value projections included in the operator.
All attention branches below reuse these block-input projections.

\paragraph{Parallel attention branches.}
In the video expert, current tokens read $C$, world-change tokens read $[C;W]$, and future-video tokens read $[C;F]$.
Thus, $W$ and $F$ form parallel prediction streams conditioned on the observed scene.
In the action expert, action tokens read the scene and world-change features alongside their own hypothesis, while effect tokens read the scene and action hypothesis alongside one another:
\begin{equation}
\begin{aligned}
    \widetilde{\mathbf A}^n&=\operatorname{Attn}\!\left(
    \mathbf A^n,[\mathbf C^n;\mathbf W^n;\mathbf A^n],[\mathbf C^n;\mathbf W^n;\mathbf A^n]
    \right),\\
    \widetilde{\mathbf E}^n&=\operatorname{Attn}\!\left(
    \mathbf E^n,[\mathbf C^n;\mathbf A^n;\mathbf E^n],[\mathbf C^n;\mathbf A^n;\mathbf E^n]
    \right).
\end{aligned}
    \label{eq:impl_action_attention}
\end{equation}

Reverse guidance and feedback use separate attention branches, contributing to the future-video and action attention outputs, respectively:
\begin{equation}
\begin{aligned}
    \widetilde{\mathbf F}^n&\leftarrow\widetilde{\mathbf F}^n+
    \operatorname{Attn}(\mathbf F^n,\mathbf E^n,\mathbf E^n),\\
    \widetilde{\mathbf A}^n&\leftarrow\widetilde{\mathbf A}^n+
    \operatorname{Attn}(\mathbf A^n,\mathbf E^n,\mathbf E^n).
\end{aligned}
    \label{eq:impl_effect_residuals}
\end{equation}
These branches also read $\mathbf E^n$, the effect features entering the block.
The updated effect features $\mathbf E^{n+1}$ become available to attention in the next block.

\paragraph{Completing the expert blocks.}
After combining its attention outputs, each expert applies its own output projection and residual connection.
The resulting features then query the instruction embeddings and projected robot-state token through conditioning cross-attention.
This conditioning memory is distinct from the visual tokens $\mathbf C$ used above.
A feed-forward network with normalization and residual connections completes each block, producing $[\mathbf C^{n+1};\mathbf W^{n+1};\mathbf F^{n+1}]$ and $[\mathbf A^{n+1};\mathbf E^{n+1}]$.
The two experts use separate parameters throughout these operations.
Prediction heads and predictive feature alignment operate on the final-layer features.

\paragraph{Action inference.}
For each observation, the video expert processes $\mathbf C$ and $\mathbf W$ once and caches their keys and values at every layer.
These groups do not attend to $F$, $A$, or $E$, so their features can be reused throughout action denoising.
At each denoising step, the action expert processes $\mathbf A$ and $\mathbf E$ through the stack, reading this cache and using effect-to-action feedback.
Future-video tokens and the training-only predictive readout are unnecessary for this path.

\subsection{Predictive Readout and Its Training}
\label{sec:appendix_predictive_readout}

We describe the three-source predictive readout in \Cref{fig:supp_predictive_readout}.

\begin{figure*}[t]
    \centering
    \includegraphics[width=\textwidth]{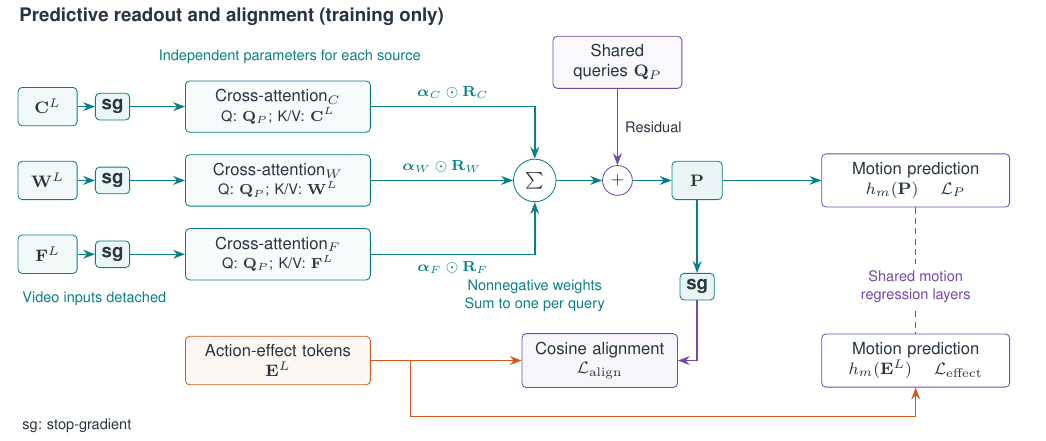}
    \caption{\textbf{Training the predictive readout.}
    A common set of queries reads the final-layer C/W/F features through three independently parameterized cross-attention branches.
    Learnable, normalized weights combine their outputs, followed by a query residual.
    Motion supervision trains the resulting predictive features $\mathbf P$, sharing the motion-regression layers with the action-effect prediction head $h_m$.
    Stop-gradients isolate the video inputs from readout supervision and make $\mathbf P$ a fixed target for each alignment update.
    }
    \label{fig:supp_predictive_readout}
\end{figure*}

\paragraph{Independent readouts and gated fusion.}
Eight learnable queries, augmented with temporal encodings, form $\mathbf Q_P$.
They read current-observation, world-change, and future-video features in parallel.
Each source $X\in\{C,W,F\}$ has an independently parameterized cross-attention branch:
\begin{equation}
    \mathbf R_X=\operatorname{CrossAttn}_X\!\left(
       \mathbf Q_P,\operatorname{sg}(\mathbf X^L),
       \operatorname{sg}(\mathbf X^L)\right).
    \label{eq:supp_source_readout}
\end{equation}
Input normalization and dimensional projections are included in the attention and prediction operators throughout this subsection.
Learnable source weights combine the readouts:
\begin{equation}
    \mathbf P=\mathbf Q_P+
    \sum_{X\in\{C,W,F\}}\boldsymbol\alpha_X\odot\mathbf R_X.
    \label{eq:supp_predictive_mixture}
\end{equation}

\paragraph{Supervision of the readout.}
The predictive features $\mathbf P$ are supervised to recover the same end-effector pose changes as the action-effect tokens $\mathbf E^L$.
Here, $h_m$ is the action-side motion prediction head introduced in the main text.
We train the predictive readout using
\begin{equation}
    \mathcal L_P=\lambda_P\,
        \ell_{\mathrm{Huber}}(h_m(\mathbf P),\mathbf m).
    \label{eq:supp_predictive_supervision}
\end{equation}
The P and E paths share the final motion-regression layers; input adaptation is absorbed into $h_m$, as in the main text.

\paragraph{Alignment and gradient flow.}
We align corresponding action and predictive features in their shared feature space:
\begin{equation}
    \mathcal L_{\mathrm{align}}=
    \operatorname{Mean}_{\mathrm{valid}}\!\left[
        1-\cos(\mathbf E^L,\operatorname{sg}(\mathbf P))\right],
    \label{eq:supp_alignment_space}
\end{equation}
As in the main text, cosine similarity is evaluated in a common feature space, with the required feature mapping left implicit.

\section{Training and Comparison Details}
\label{sec:appendix_impl}

\subsection{Target Construction}
\label{sec:appendix_targets}
\paragraph{Video and depth targets.}
Appearance-change targets are differences between adjacent clean VAE latents; their temporal grid can differ from the sampled RGB/depth grid because the VAE compresses time.
Depth changes use adjacent sampled frames from the selected camera view and current-frame median-depth normalization, with confidence and padding masks.

\paragraph{Robot-motion targets.}
Following the main text, the motion target at future offset $\delta_i$ comprises end-effector translation and rotation changes:
\begin{equation}
    \mathbf m_i=[\Delta\mathbf p_i;\Delta\mathbf r_i],
    \qquad \Delta\mathbf p_i=\mathbf p_{t+\delta_i}-\mathbf p_t.
    \label{eq:motion_target}
\end{equation}
Here, $\mathbf p_t$ is the end-effector position in world coordinates, and $\Delta\mathbf r_i$ represents its relative rotation as an axis--angle vector.
LIBERO targets use the recorded end-effector states.
For RoboTwin, these are computed from the joint-command sequence using a dual-arm kinematic proxy, rather than measured end-effector states.
Pose channels are standardized using valid training-set targets, and physical readout errors are computed after reversing this normalization.

\paragraph{Validity masks.}
An offset contributes only when the current state and images, the required action prefix, and the corresponding future image and target are valid.
Pose and gripper regression use their respective validity masks; feature alignment uses their intersection.
Regression losses average valid coordinates within each sample and then average samples; masked samples contribute zero.
These masks also exclude temporal padding from the diagnostic metrics.

\subsection{Loss Weights}
\label{sec:appendix_losses}
\label{sec:appendix_ablation}

Tab.~\ref{tab:weights} lists the coefficients of $\mathcal L_{\mathrm{change}}$, $\mathcal L_{\mathrm{effect}}$, $\mathcal L_{\mathrm{align}}$, and $\mathcal L_P$ in the Fast-WAM-based implementation.

\begin{table}[t]
\centering\small
\setlength{\tabcolsep}{3pt}
\caption{\textbf{Weights for change prediction, feature alignment, and predictive-readout supervision.} The first three rows form $\mathcal L_{\mathrm{change}}$; the remaining rows correspond to $\mathcal L_{\mathrm{effect}}$, $\mathcal L_{\mathrm{align}}$, and $\mathcal L_P$.}
\label{tab:weights}
\begin{tabular}{lcr}
\toprule Objective & Coefficient & Weight \\
\midrule
Video-latent difference & $\lambda_z$ & 0.5 \\
Depth difference & $\lambda_d$ & 0.1 \\
W motion prediction & $\lambda_m^v$ & 0.1 \\
E motion prediction & $\lambda_m^a$ & 0.04 \\
Feature alignment & $\lambda_{\mathrm{align}}$ & 0.02 \\
P motion prediction & $\lambda_P$ & 0.04 \\
\bottomrule
\end{tabular}
\end{table}

The original flow-matching losses retain Fast-WAM's scheduler-dependent weighting and action coefficient 1.0.

\subsection{Related-Method Implementations}
\label{sec:appendix_comparison}
\label{sec:appendix_addon}

\paragraph{Common-backbone adaptations.}
For Track4Action and DreamWAM, we implement their enhancement mechanisms on the common Fast-WAM backbone, retaining the  matched downstream training budgets.
The Track4Action adaptation uses 16 queries of width 512, two attention layers, and eight heads to read current-frame features at block 15.
It distills a frozen, pooled Track4World descriptor for the 32-step window using masked MSE with weight 0.1; a zero-initialized feature-wise gate fuses these query features into the visual features read by the action expert.
The descriptor construction and injection configuration are adapted to this backbone.

The DreamWAM adaptation retains RGB/optical-flow joint denoising, DINOv2 and depth prediction, and the released residual-routing design.
Its rank-eight target projections are fitted on 128 fixed training windows, without evaluation data, and it retains the released RGB/flow output-head initialization.

\begin{table*}[t]
\centering\small
\setlength{\tabcolsep}{6pt}
\caption{\textbf{LIBERO-Plus results across suites.} Our evaluations cover all 10,030 cases, with Avg. equally averaging the four suite success rates (\%).
$^\dagger$ denotes improved initialization.}
\label{tab:plus_details}
\begin{tabular}{@{}llrrrrr@{}}
\toprule
Base WAM & Variant & Spatial & Object & Goal & Long & Avg. \\
\midrule
\multirow{2}{*}{Fast-WAM} & Base & 52.21  & 69.06  & 37.32  & 40.06  & 49.66 \\
 & +Ours & \textbf{87.34} & \textbf{89.41} & \textbf{66.22} & \textbf{73.12} & \textbf{79.02} \\
\midrule
\multirow{2}{*}{4D-WAM} & Base & 79.43 & 74.70 & 58.05 & 63.83 & 69.00 \\
 & +Ours & \textbf{88.51} & \textbf{88.72} & \textbf{69.93} & \textbf{73.36} & \textbf{80.13} \\
\midrule
\multirow{2}{*}{4D-WAM$^\dagger$} & Base & 86.84 & 81.65 & 74.68 & 78.29 & 80.37 \\
 & +Ours & \textbf{90.63} & \textbf{89.28} & \textbf{75.45} & \textbf{83.37} & \textbf{84.68} \\
\midrule
\multirow{2}{*}{InternW0-$\Delta$} & Base & 90.05 & 97.10 & 88.31 & \textbf{95.67} & 92.78 \\
 & +Ours & \textbf{91.55} & \textbf{97.93} & \textbf{91.43} & 94.80 & \textbf{93.93} \\
\bottomrule
\end{tabular}
\end{table*}

\section{Additional Analyses and Experiments}
\label{sec:appendix_analysis}

\subsection{Detailed LIBERO-Plus Results}
\label{sec:appendix_policy_results}

\paragraph{LIBERO-Plus.}
\Cref{tab:plus_details} expands the aggregate results in Table 1 of the main paper into the Spatial, Object, Goal, and Long suites.
The two 4D-WAM comparisons use the same architecture and matched training budgets within each initialization setting.
Incorporating our proposed mechanisms improves all four suites under both settings.

\subsection{Gains at Different Baseline Performance}
\label{sec:appendix_gain_scaling}

\begin{table}[t]
\centering\small
\setlength{\tabcolsep}{3pt}
\caption{\textbf{Add-on gains at different baseline success rates.} LIBERO-Plus results using equal suite averaging.
Base is success (\%); Gain is in percentage points.
Failure red. is the relative reduction in average failure rate, computed as $100\times\mathrm{Gain}/(100-\mathrm{Base})$.
$^\dagger$ denotes improved initialization.}
\label{tab:gain_scaling}
\begin{tabular}{@{}lrrr@{}}
\toprule
Base WAM & Base & Gain & Failure red. (\%) \\
\midrule
4D-WAM & 69.00 & +11.13 & 35.90 \\
4D-WAM$^\dagger$ & 80.37 & +4.32 & 21.99 \\
InternW0-$\Delta$ & 92.78 & +1.15 & 15.88 \\
\bottomrule
\end{tabular}
\end{table}

The reported LIBERO-Plus comparisons show smaller absolute gains at higher baseline success rates (\Cref{tab:gain_scaling}).
For the same 4D-WAM architecture, improved initialization raises baseline success from 69.00\% to 80.37\%, while the add-on gain decreases from 11.13 to 4.32 percentage points.
InternW0-$\Delta$ comparison starts at 92.78\% and gains 1.15 points.
A higher baseline leaves fewer failures to correct.
To account for this, \Cref{tab:gain_scaling} also reports the relative reduction in average failure rate, computed as the success-rate gain divided by $100$ minus baseline success.
For the two 4D-WAM settings, this reduction is 35.90\% and 21.99\%, respectively.

\paragraph{Which errors remain matters.}
The perturbation categories in Table 2 of the main paper help distinguish these effects without relying on individual task outcomes.
For background changes, the gain drops from 16.36 to 4.93 points, but the relative failure-rate reduction remains similar: 40.18\% versus 37.86\%.
Here, the higher starting success rate accounts for much of the smaller absolute gain.
Camera changes show a different pattern: baseline success remains moderate, increasing from 42.96\% to 50.84\%, while the relative failure-rate reduction falls from 37.39\% to 20.36\%.
For sensor noise, it falls from 58.79\% to 27.33\%.
These observations suggest that the remaining types of errors, as well as the starting success rate, affect the benefit of change-based guidance.
Overlap with knowledge already provided by initialization is one possible explanation.

\subsection{Action Dependence of Video Prediction}
\label{sec:appendix_action_conditioning}

Using the same bidirectional model as in Section 4.5 of the main paper, we find that the reverse pathway makes future-video predictions responsive to the supplied actions.
Across 40 LIBERO tasks and two starting states per task, we execute two 32-step sequences from each shared state: the recorded actions and a version with translation and rotation commands halved, leaving gripper commands unchanged.
During video generation, we vary only the supplied action sequence, keeping current images, instruction, robot state, and sampling noise fixed; no future images are provided.
With two generation seeds, disabling reverse guidance yields bitwise-identical video predictions in all 160 paired comparisons, whereas enabling it changes the predicted future latents in every comparison.
At the final prediction step (step 32), the RGB RMSE between the frames generated under the two action sequences is 6.75 with reverse guidance and exactly zero without it.
RMSE is computed after averaging MSE over both cameras, seeds, and starting states within each task, then equally across tasks.

\fi
\end{document}